\documentclass[letterpaper, 10 pt, conference]{ieeeconf}  

\IEEEoverridecommandlockouts                              

\usepackage{amsmath} 
\usepackage{amssymb}  
\usepackage{graphicx}
\usepackage{subcaption}

\DeclareMathOperator*{\argmax}{arg\,max}

\newtheorem{remark}{Remark}

\title{\LARGE \bf
Learning to Explore in Motion and Interaction Tasks
}

\author{Miroslav Bogdanovic$^{1}$ and Ludovic Righetti$^{1, 2}$
\thanks{$^{1}$Movement Generation and Control group, Max-Planck Institute  for  Intelligent  Systems,  T\"ubingen,  Germany.  Email:  ̈first.lastname@tuebingen.mpg.de}%
\thanks{$^{2}$Tandon  School  of  Engineering,  New  York  University,  USA.}%
\thanks{This work was supported by New York University, the Max-Planck Society and the European Unions Horizon 2020 research and innovation program (grant agreement No 780684 and European Research Councils grant No 637935).}
}

\begin{document}

\maketitle
\thispagestyle{empty}
\pagestyle{empty}

\begin{abstract}
Model free reinforcement learning suffers from the high sampling
complexity inherent to robotic manipulation or locomotion tasks. Most successful approaches typically use random sampling strategies which leads to slow policy convergence.
In this paper we present a novel approach for efficient exploration that leverages previously learned tasks. We exploit the fact that the same system is used across many tasks and build a generative model for exploration based on data from previously solved tasks to improve learning new tasks. The approach also enables continuous learning of improved exploration strategies as novel tasks are learned.
Extensive simulations on a robot manipulator performing a variety of motion and contact interaction tasks demonstrate the capabilities of the approach. In particular, our experiments suggest that the exploration strategy can more than double learning speed, especially when rewards are sparse. Moreover, the algorithm is robust to task variations and parameter tuning, making it beneficial for complex robotic problems.

\end{abstract}

\section{INTRODUCTION}

Deep reinforcement learning has attracted a lot of attention for robotic applications where full robot models can be difficult to identify, especially for contact dynamics, and lead to computationally challenging planning and control problems. In particular, it can be successful in producing robust behaviors with ability to handle uncertainties in the environment and quickly adapt to changes \cite{hwangbo2019learning}.

However, these algorithms suffer from several important issues
that can limit their applicability. The most salient issue is
related to sampling efficiency and exploration strategies.
Indeed, exploration strategies required to generate new samples are often limited to simple noise models, which can drastically increase the number of required samples for policy convergence.
This problem is especially acute in robotics, due to difficulties in obtaining large amounts of training data, discontinuities in the interactions with the environment, as well as complex, multi-part, potentially sparse reward functions. Quite often algorithms suffer from local minimums and flat surfaces in the reward space.
In  this  work,  we  investigate  a  novel  exploration  strategy  to
alleviate  these  issues by leveraging previously learned tasks to better explore when learning novel ones.

One attempt to tackle the issue of exploration when applying deep reinforcement learning in robotics problems has been made in \cite{wawrzynski2015control}, by proposing the application of correlated noise. Correlated noise, more specifically Ornstein-Uhlenbeck (OU) process \cite{PhysRev.36.823}, is also used to improve exploration in \cite{DBLP:journals/corr/LillicrapHPHETS15}. While this exploration strategy can in theory explore the state space more rapidly, it tends to create very high changes in the control sequence while it would be preferable in robotic applications to have smoother motions with proper velocity profiles.

Exploration in reinforcement learning has been explored in more general settings. \cite{DBLP:journals/corr/PlappertHDSCCAA17}, for example, proposes exploration in the space of the parameters of the policy instead of the space of actions. Other approaches base exploration on some measure of novelty while moving through the state space, as is the case in research on intrinsic motivation \cite{Barto2004, laversanne2018curiosity, Pathak2017}.
These approaches mostly try to explore novel regions of the state space, but do not necessarily use knowledge from previously learned tasks. In this paper, we take a complementary approach where we leverage these previous tasks to generate better exploration strategies for novel ones.

Transfer learning, which seeks to exploit knowledge from previously learned tasks to accelerate learning new tasks is also relevant to our problem. 
In \cite{DBLP:journals/corr/TehBCQKHHP17}, authors propose learning a "distilled" policy from several tasks capturing the common behavior among them and constraining the individual policies to be close to it. The exploration process we learn bears some resemblance to this shared policy, the key difference being that in continuous action spaces, which is the setup we investigate, having such distribution only be dependent on the current state does not prove informative enough as we explain in detail in Section \ref{sec:toy_example}.
In \cite{DBLP:journals/corr/HeessWTLRS16}, low level control on a locomotion system is learned and used to solve high level tasks. However, it requires that this type of separation exist as well as enough knowledge about it to be able to design a two-level structure with capability of learning it.
Auxiliary tasks can also be used to improve learning when rewards
are sparse, as in \cite{DBLP:journals/corr/abs-1802-10567}, however this requires an ad-hoc setup to define these tasks and how they should be interleaved with the main task to be learned. Our approach does not setup auxiliary tasks but still assumes that a set of tasks of increasing complexity is available.
Our work also bears some connection to research on motion primitives \cite{schaal2006dynamic}, \cite{ijspeert2013dynamical}. While we do not explicitly try to directly extract any motion primitive, the exploration process we learn can be thought of as generating basic action primitives.

In this work, we present a novel exploration strategy for deep reinforcement learning. We propose to learn a generative model
of basic action primitives capturing the motion patterns seen in previously learned tasks. Extensive experiments on a set of
simulated motion and contact interaction tasks for a robot manipulator demonstrate the capabilities of the approach. In particular, our approach shows significant learning speed-up compared to other state of the art algorithms, especially for tasks with sparse reward. We also show that the algorithm is robust to parameter changes and task variations, reducing the need for parameter tuning.

\section{PROPOSED APPROACH}

Our goal in this work is to use the knowledge we gain in solving several tasks with a given robot to facilitate learning new tasks for the same robot. In all the tasks we discuss in the paper we use
the same robot and vary the tasks and the environments, with increasing complexity. More specifically, we want a method that uses good behaviors learned from previous tasks to help the exploration process when learning a new task. Our idea is to learn a function
that generates random behaviors that resemble behaviors
seen in previous tasks.
In the following, we first describe our approach to build a model that generates random behaviors that retain similar characteristics as the behaviors learned in the previous tasks.

\begin{figure*}
\includegraphics[width=\linewidth]{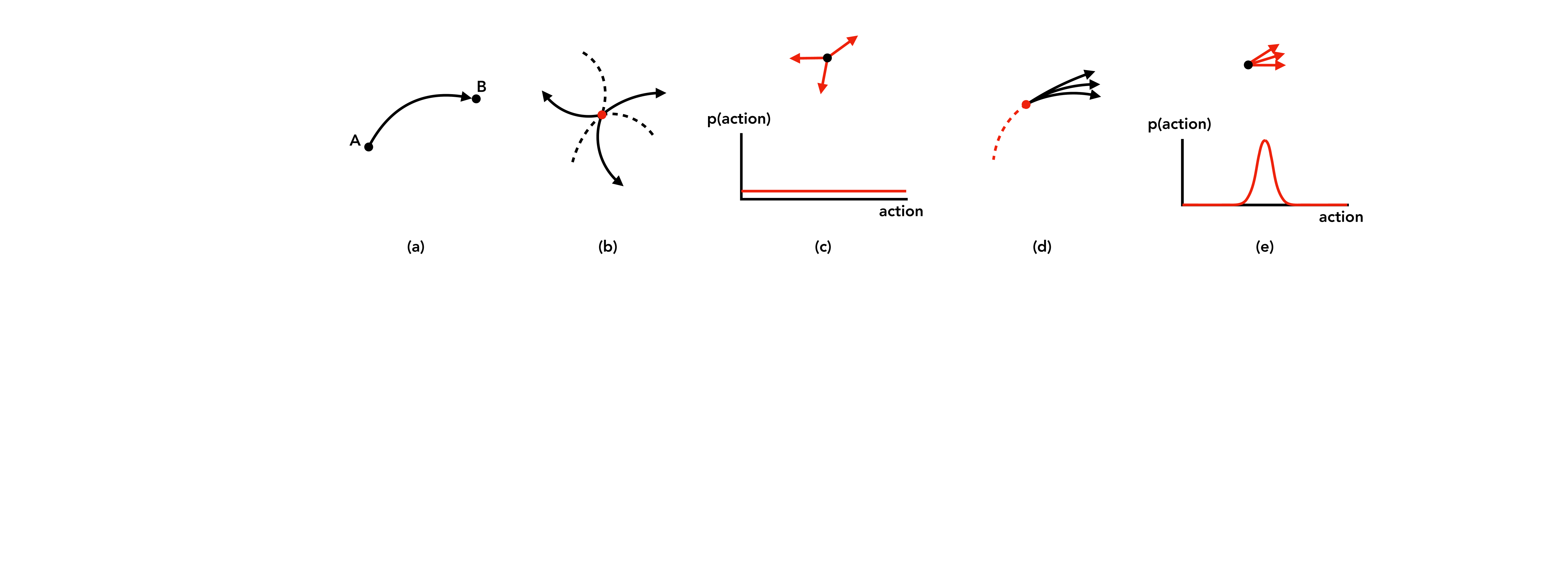}
\centering
\caption{Point mass example illustrating our approach (cf. Section \ref{sec:toy_example} for details): (a) Task: moving from one point to another; (b) Policy trajectories conditioned on the selected state; (c) Combined action distribution conditioned on the selected state; (d) Policy trajectories conditioned on a state history; (e) Combined action distribution conditioned on a state history.}
\label{fig:example_env}
\end{figure*}

\subsection{Learning the exploration model}\label{sec:toy_example}
We start by presenting a simple example illustrating the requirements
for our approach and the difficulties inherent to learning good exploration strategies from previous tasks. In particular, we 
want to explain why longer trajectories need to be taken into
account to learn proper exploration strategies.
We consider a point mass moving in the plane with a fixed velocity
magnitude. The control input consists in choosing a direction of motion, independently at each state. The desired task is to reach a desired location (B) from a randomly selected starting point in the plane (A) as shown in Figure \ref{fig:example_env}(a).

We collect successful policies $\pi_i$ for many instances of such tasks (i.e. for different goal positions). We then choose an arbitrary state $s$ and evaluate the actions that all the different policies would take at that state $a_i = \pi_i(s)$ (Figure \ref{fig:example_env}(b)). Considering that for each policy, the random goal positions can be distributed anywhere around the chosen state $s$, we can expect that given enough policies, we will be
able to find a policy that would move in any chosen direction.
Therefore, if we were to calculate a combined distribution $\pi_c$ over the action space arising from all the policies:

\begin{equation}
    \pi_c(a \mid s) \approx \sum_i \pi_i(a \mid s),
\end{equation}

we would likely find a uniform distribution (i.e. no preferred direction of motion) which would be completely uninformative (Figure \ref{fig:example_env}(c)). This illustrates that we cannot naively 
combine previous actions at a given state because the resulting distribution is likely to be of little interest to create sensible exploratory motions.

However, if we condition such distribution on a longer history of preceding states $s_{1:t}$, then there will exist only a few policies with a similar state history when arriving at the state (Figure \ref{fig:example_env}(d)) and it is likely that their subsequent actions would be very similar. The combined probability distribution over actions:

\begin{equation}
    \pi_c(a \mid s_{1:t}) \approx \sum_i p(s_{1:t} \mid \pi_i) \pi_i(a \mid s_{1:t}),
\end{equation}

where $p(s_{1:t} \mid \pi_i)$ is the weighting factor equal to the probability of the policy $\pi_i$ resulting in a state trajectory $s_{1:t}$, will be much more focused (Figure \ref{fig:example_env}(e)). Taking consecutive samples from such a distribution would now result in a behavior with the characteristics of the original policies and would be significantly more useful for what we are trying to achieve.

We propose in the following to learn a combined probability distribution of a diverse set of policies conditioned on trajectories (or sequences) of preceding states.
To represent this learned exploration model (LEP), we use a recurrent network, more specifically a Long Short Term Memory (LSTM, \cite{hochreiter1997long}) network. We start by collecting trajectories $\{(s_{1:T}, a_{1:T})_i\}$ from all the available policies. We note here that we only need trajectories and not the full policies for training the model, so our approach could work even if that is all we have access to (for example as a result of doing trajectory optimization \cite{Mordatch:2012wm, ponton_time_2018} or learning by demonstration \cite{billard2008robot}).

Importantly, we do not train our model on the full length of the collected trajectories, as we are interested in general basic characteristics of good behaviors, not behaviors that solve specific tasks. On longer time scales the characteristic associated to solving a specific task would become dominant, leading our network to overfit to solutions to individual tasks. Because of that we limit trajectory samples to a shorter timescale $h$ and randomly sample sections of the trajectories of length $h$ to build the training dataset.

The function approximated by the LEP network takes
as input a history of robot states and outputs a distribution
over the corresponding sequence of actions. We use diagonal Gaussian distribution as the output and train the network to maximize the log likelihood of the action sequences given the corresponding states (in the same way as in, for example, \cite{DBLP:journals/corr/Graves13}).

\subsection{Reinforcement learning with an exploration function}
We now describe how the exploration model can be included in
a reinforcement learning algorithm. In our experiments, we use
Deep Deterministic Policy Gradient (DDPG, \cite{DBLP:journals/corr/LillicrapHPHETS15}),
 but any off-policy algorithm with independent noise could be used instead.
DDPG is an actor-critic method that simultaneously learns a state-action value function $Q(s, a)$ and a deterministic policy $\pi(s)$ that optimizes it:

\begin{equation}
    \pi(s) = \argmax_a Q(s,a)
\end{equation}

By keeping a constant estimate of the action with the largest $Q$ value for each state it avoids the problem that arises in continuous action spaces, where calculating this value online requires solving an optimization problem.

The full algorithm consists of interchanging steps of:

\begin{enumerate}
    \item Gathering data by executing the current policy in the environment with an added output of an external noise process $\mathcal{N}$:
    \begin{equation}
        \begin{split}
            a_t &= \pi(s_t) + \epsilon \\
            \epsilon &\sim \mathcal{N}
        \end{split}
        \label{ddpg_exp}
    \end{equation}
    \item Taking random samples from the gathered data and updating the value of the $Q$ function using the Bellman equation and the policy $\pi$ corresponding to the gradient of the current $Q$ function estimate with respect to the action.
\end{enumerate}

As we have seen, DDPG explores the spaces by adding exploration noise to an existing deterministic policy (Equation \ref{ddpg_exp}).
We replace the exploration noise in DDPG 
with the output of the LEP (a generative model for motions with good properties). With that, the action that is taken at each time step during training is equal to the sum of current output of the deterministic policy and a sample from the exploration model:
\begin{equation}
    a_t = \pi(s_t) + \epsilon_{LEP}
\end{equation}

We reset the internal state of the LSTM network to its initial value every $h$ steps, matching the sequence length it has been trained on. Thereby, the action distribution of the exploration is conditioned on the past $t \mod h$ states, where t is the current time step:
\begin{equation}
    \begin{split}
        \epsilon_{LEP} &\sim p_{LEP}(a \mid s_{t_r:t}),\\
        t_r &= t - t \bmod h
    \end{split}
\end{equation}

Apart from this, we keep all the other aspects of the training exactly as they are in \cite{DBLP:journals/corr/LillicrapHPHETS15}, including not reducing the exploration noise as the training progresses.

\begin{remark}
Note that while our model might produce sensible behaviors for the system there is no guarantee that the sum with output of the current policy will do the same. However, this simple approach works
very well in practice. 
The DDPG policy is initialized to produce output values close to zero at the beginning of the training. Because of that, the initial samples in our case will for the most part be pure samples from the generative exploration model.
As the policy changes during learning this might no longer be the case, but as we will later see in our experiments, the algorithm shows no trouble finely converging to the desired behavior.
\end{remark}

\subsection{Continuous learning} \label{continous_learning}

One more aspect of our approach is its ability to be continuously applied, enabling solving of more and more complex tasks each time.
As we have stated, we can use data from any source in training of the exploration model.
This also includes policies resulting from applications of some previously trained exploration model of the same type.
This way we can train the model on the data available to us initially, use it to learn on a new task, add the data from these new policies to our training set and then repeat the process.
We can keep doing this as many times as necessary to get to a point where we can solve some complex task we are interested in.

\section{EXPERIMENTAL SETUP}

In this section we describe the experimental setup used to evaluate the performance of the approach.

\begin{figure}
\captionsetup[subfigure]{labelformat=empty}
\begin{subfigure}[b]{.242\linewidth}
        \includegraphics[width=\textwidth]{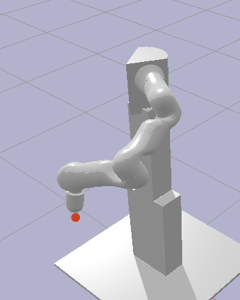}
        \caption{Task 1}
    \end{subfigure}
    \begin{subfigure}[b]{.242\linewidth}
        \includegraphics[width=\textwidth]{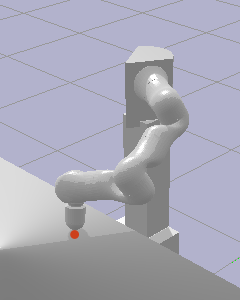}
        \caption{Task 2}
    \end{subfigure}
    \begin{subfigure}[b]{.242\linewidth}
        \includegraphics[width=\textwidth]{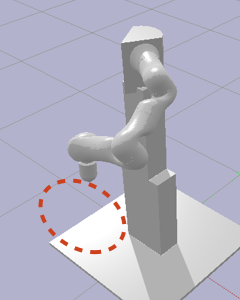}
        \caption{Task 3}
    \end{subfigure}
    \begin{subfigure}[b]{.242\linewidth}
        \includegraphics[width=\textwidth]{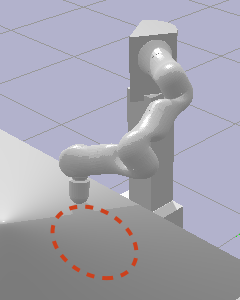}
        \caption{Task 4}
    \end{subfigure}
    \caption{Illustration of the four tasks tested in the experiments.}
\end{figure}

\subsection{Environment}
We test the algorithm on a simulated KUKA LWR,
a 7-DoF robotic arm. We choose tasks of increasing complexity and
in particular tasks involving contact interactions with a table. All our experiments are implemented using the Bullet physics simulator.
We keep the state and action spaces same across all the tasks. The state space consist of joint position, joint velocities and measured 3D contact forces at the end-effector, for a total of 17 dimensions. The action space consists of torques applied to each joint, which is 7 control dimensions. We also add gravity compensation to the torque command as it is automatically added by the on-board controllers on the real KUKA LWR robot.

\subsection{Motion and contact interaction tasks}
Each of our different tasks is characterized completely by the defined reward function (and the presence or absence of the table in the environment, which we add in tasks with interaction aspects). We test four types of tasks of increasing
complexity: a reaching task, a contact task, a periodic motion
and a periodic motion while interacting with the table.
We describe the tasks and the cost functions in detail in the following. 
Note that our cost functions are rather straightforward and do not specially seek to facilitate learning.

\subsubsection{Task 1: Reaching a desired target}

This task consists of getting the end-effector to a desired position and orientation in space. The reward function consists of two parts:
the distance to the goal and the orientation error.
In this task, the desired orientation always points straight down and we only vary the desired goal position.

\subsubsection{Task 2: Stationary force application}

The goal of this task is to apply a desired normal force on a desired location on the table. The reward function in this case will consists of three parts, the same two costs used in Task 1 for the position and orientation of the end-effector and a cost measuring the error between the desired and measured normal contact forces and adding
a constant bonus term whenever the end-effector is in contact with the table to incentivize contact behaviors. Note that the contact cost
is sparse as most robot configurations lead to no contacts.

\subsubsection{Task 3: Periodic motion along a closed curve}

Here we require the end-effector to move along a given circular path in space (while keeping a specified orientation). The reward function has three parts, based on position, velocity and orientation of the end-effector. The reward is based on the distance to the circle (as opposed to a fixed target location as in Task 1) and desired velocity based on the tangential velocity vector that the end-effector should have on the point on the circle currently closest to it. The orientation reward is kept the same as in previous tasks.
 
\subsubsection{Task 4: Periodic motion with contact force regulation}

Combining aspects of all the previous tasks, here the goal is to move along a given circular trajectory on the surface of the table while applying a constant normal force to it. Reward function is a combination of the trajectory reward given in Task 3 and the force reward used in Task 2.

\subsection{Defining success during learning}
It is not sufficient to find a high reward policy, we also
need to check that the robot is indeed achieving the desired task.
For each task, we empirically define a reward value for which
we consider the task solved. In order to find this value, we analyzed many instances of the behavior on the task. We determine the value such that all behaviors with higher scores perform all the aspects of the task in a satisfactory way. For example, in Task 4, we make sure that policies performing only three out of the four aspects of the behavior we desire (moving along the trajectory without making contact with the table, applying force while being stationary on a single point on a trajectory, etc.) never reach this threshold for the cumulative reward.

\section{Results}
We now present the results of our simulations.
In particular, we demonstrate how tasks involving complex
contact interactions can be learned efficiently with our approach.
We also systematically compare the results with other state
of the art reinforcement learning algorithms and test the robustness of the approach to random initialization.
In all of our experiment, we compare our method with the normal DDPG algorithm and with an on-policy reinforcement algorithm, Proximal Policy Optimization (PPO) \cite{DBLP:journals/corr/SchulmanWDRK17}.
For both these algorithms, we use the implementations from OpenAI Baselines \cite{baselines}.

\subsection{Same task type for training and testing}

First, we evaluate how our method performs when it is trained on the same type of tasks it is later tested on. We investigate this 
behavior with the reaching task (Task 1), which is the simplest of all tasks.
To generate the initial policies to train the exploration model we use PPO. The reason to use this algorithm instead of DDPG is that 
the vanilla DDPG (e.g. without a good exploration strategy) did not produce policies as good as PPO for this simple experiment. All subsequent training and improvement of the exploration model are done
using data generated by our approach on previous tasks.
We use 100 policies trained on instances of the task with varying goal positions for this initial data collection.

\begin{figure}
\includegraphics[width=\linewidth]{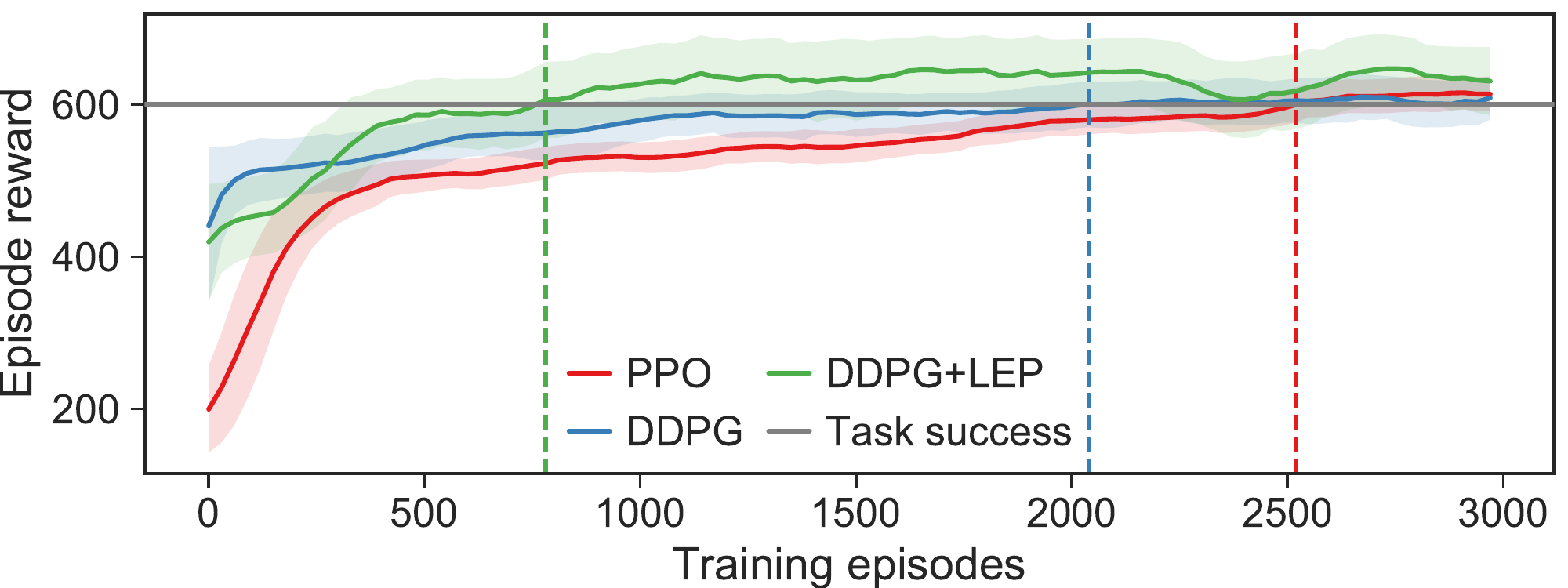}
\centering
\caption{Learning results of the best performing policies for Task 1: average cumulative reward (bold lines) and variance (shaded area) across all task instances. Our method (green) converges more than twice faster than the other algorithms (vertical dashed lines).}
\label{fig:exp1}
\end{figure}

We generate new instances of the same task with different goal locations and compare the performance of our approach (DDPG + LEP) with DDPG and PPO. We explore various noise setups for DDPG (Gaussian noise and Ornstein-Uhlenbeck processes, with a range of values for the variance), as well as different subsequence lengths for training of the LEP. We present results for the best configuration for each algorithm averaged over 6 different goal positions in Figure \ref{fig:exp1}.

We notice that all the algorithms converge to a good behavior (above the gray line), which is expected for this simple task.
PPO, being an on-policy algorithm, converges noticeably slower than the other two.
While standard DDPG implementation and the one using our exploration process both converge relatively quickly, the average for our approach reaches the desired task value more than two times faster than the one using standard noise and has a visibly higher percentage of satisfactory solutions at the end. 
While for this task the comparison can be biased as our RNN was trained on other instances of the same task (i.e. other desired positions), it is important to notice that our approach does not exhibit bias towards goal positions seen in the training set and manages to converge to the desired target with as good or better accuracy than the other two methods.

\subsection{Application to a new task and dealing with sparse reward}

\begin{figure}
\includegraphics[width=\linewidth]{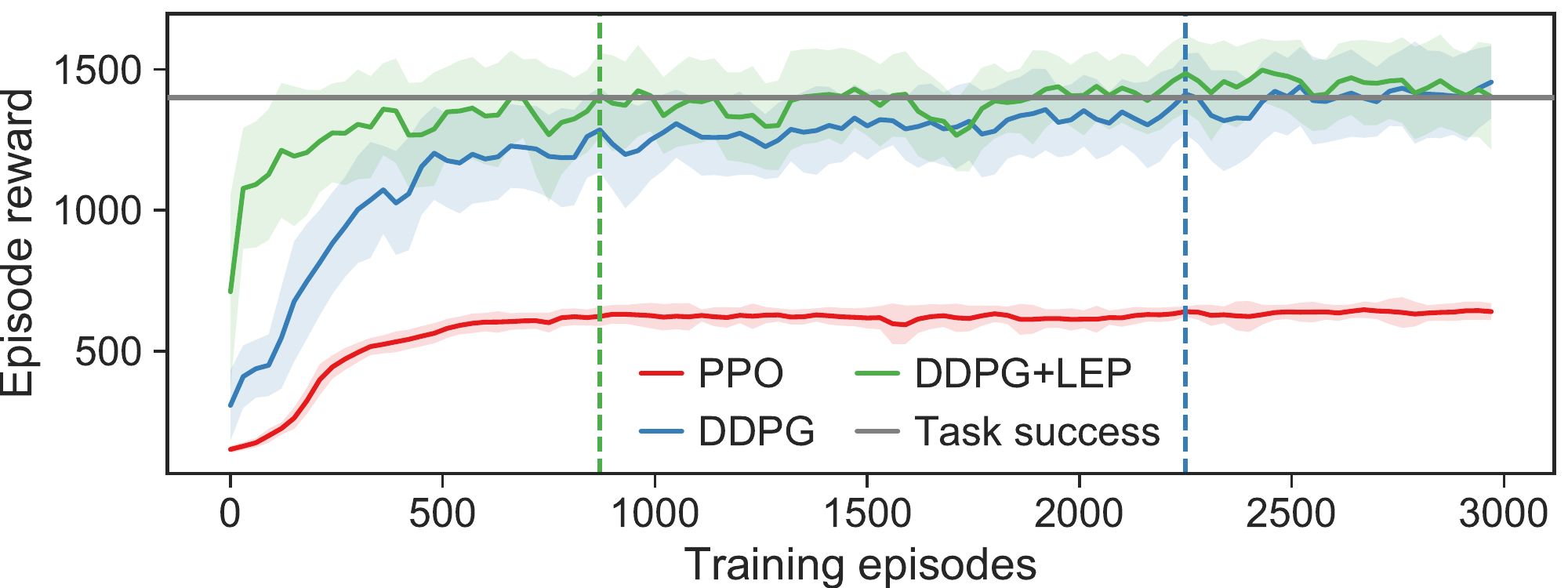}
\centering
\caption{Learning results of the best performing policies for Task 2: average cumulative reward (bold lines) and variance (shaded area) across all task instances. Our method (green) converges more than twice faster than DDPG (vertical dashed lines), PPO is not able to find a solution.}
\label{fig:exp2}
\end{figure}

Next, we investigate how our approach performs on a previously unseen task.
We are also interested in how the approach deals with a sparse reward signal. We first use Task 2 with the same exploration process trained in the previous experiment, using data from policies reaching random points in space.
We expect that our exploration model contains more informative motions for the end-effector resulting in a speedup in learning
despite the very different nature of the tasks the RNN was trained on. Indeed, this task contains a reaching component which was seen before and a contact regulation component that was not seen in Task 1.
As before, we compare our approach with regular DDPG and PPO.
The results, again averaged over 6 different task instances, are shown in Figure \ref{fig:exp2}.

Unlike the previous set of experiments, in this task PPO is not
capable of finding a policy that achieves the desired behavior.
The reward information about force interaction in this case is very sparse and is only present when the end-effector is in contact with the table.
Even though the position part of the reward guides the policy to such states, a broader exploration is then required to find rewarding types of interaction.
PPO, lacking this, fails to find solutions for the task and optimizes only for the position and orientation parts of the reward.
The other two approaches both converge to satisfactory behaviors, but again our methods does so more than twice as fast as the normal DDPG algorithm.

\begin{figure}
\includegraphics[width=\linewidth]{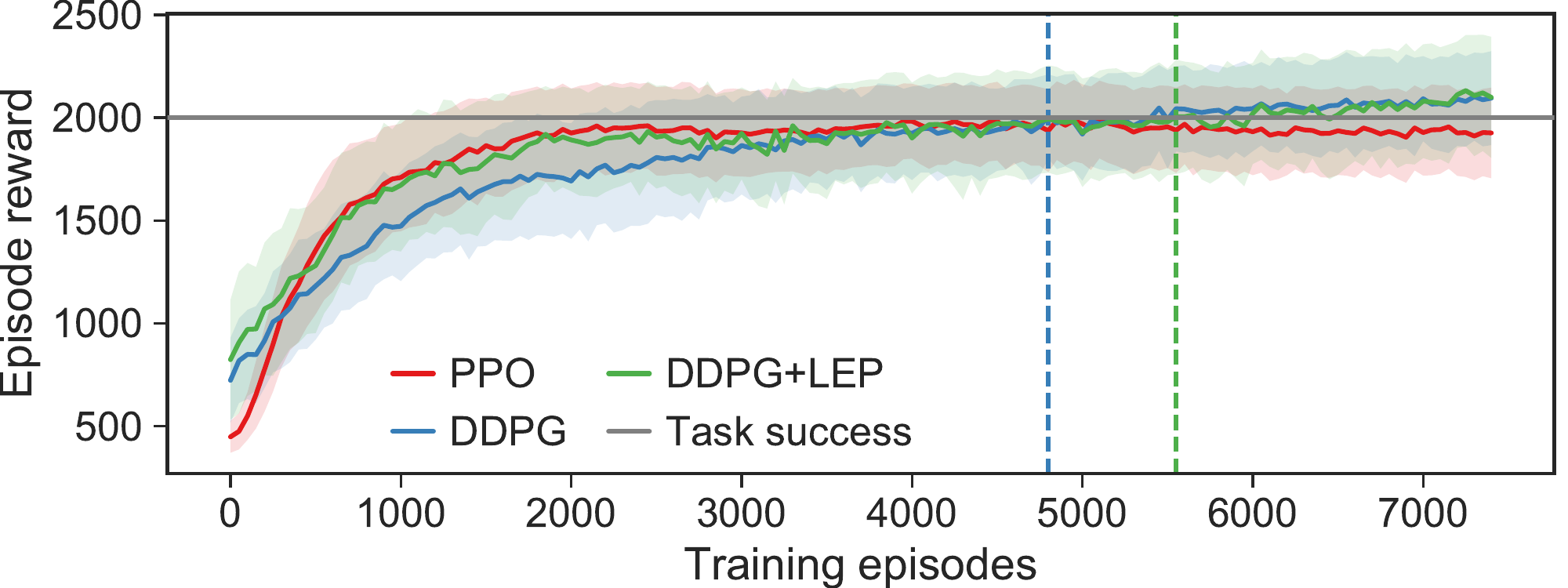}
\centering
\caption{Learning results of the best performing policies for Task 3: average cumulative reward (bold lines) and variance (shaded area) across all task instances. Our method (green) converges slightly slower that the normal DDPG (vertical dashed lines), PPO is not able to find a solution.}
\label{fig:exp_circle}
\end{figure}

We repeat this experiment with Task 3 requiring a motion along a closed curve in free space. This task requires a periodic motion, which is inherently different than reaching motions in terms of expected position and velocity profiles.
It is therefore an inherently more difficult task for our exploration model, but one with a non-sparse reward, providing guiding information throughout the state space. Here we vary the circle radius and center for each new task instance.
Results, in this case averaged over 18 different instances, are shown in Figure \ref{fig:exp_circle}.

Our approach and DDPG take a significantly longer time to converge than in the previous tasks. While PPO never reaches the threshold for which we consider that the appropriate behavior is achieved, its final behavior is not very far from being acceptable.
We notice that all three algorithms converge similarly, with the
normal DDPG being initially slower. 
This experiment suggests that for tasks that require
very different movement profiles than the movements our exploration model was trained on, our exploration method will not necessarily significantly improve convergence, yet it is still not detrimental to the learning process, which  is an important aspect to afford generalization to other tasks.

From these two experiments we can see that, as expected, the advantage for using our approach comes when the reward information is sparse and simple exploration is no longer sufficient, however the required basic movement profiles need to share similar characteristics in order to benefit from a significant speedup.

\subsection{Complex interaction task and continuous learning}

In the last experiment, we would like to demonstrate other important
aspects of our approach: that it can scale to significantly more
complex tasks and that the exploration model can be extended with previously learned motions, allowing continuous learning of richer exploration strategies as we discussed in Section \ref{continous_learning}. To do so, we update our exploration model by training it with a combination of data from Task 2 and Task 3. We collect 100 policies from each of the two tasks.

For testing we use Task 4, which requires concurrent force regulation and motion along a circular path on the table and contains aspects from all previously encountered tasks. It means that we could decompose
the task as a combination of all previous three tasks (Task 1 to reach the table, and Task 2 and 3 to perform the motion on the table).
\begin{figure}
\centering
        \includegraphics[width=\linewidth]{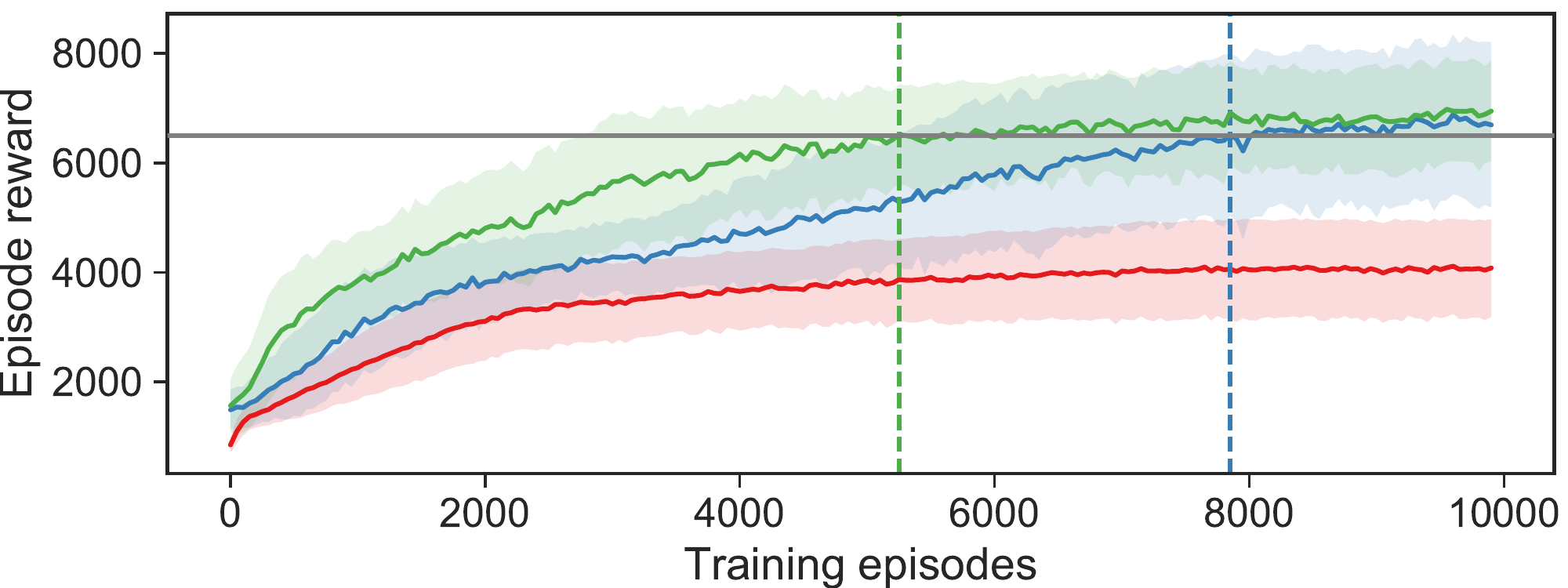}
        \includegraphics[width=\linewidth]{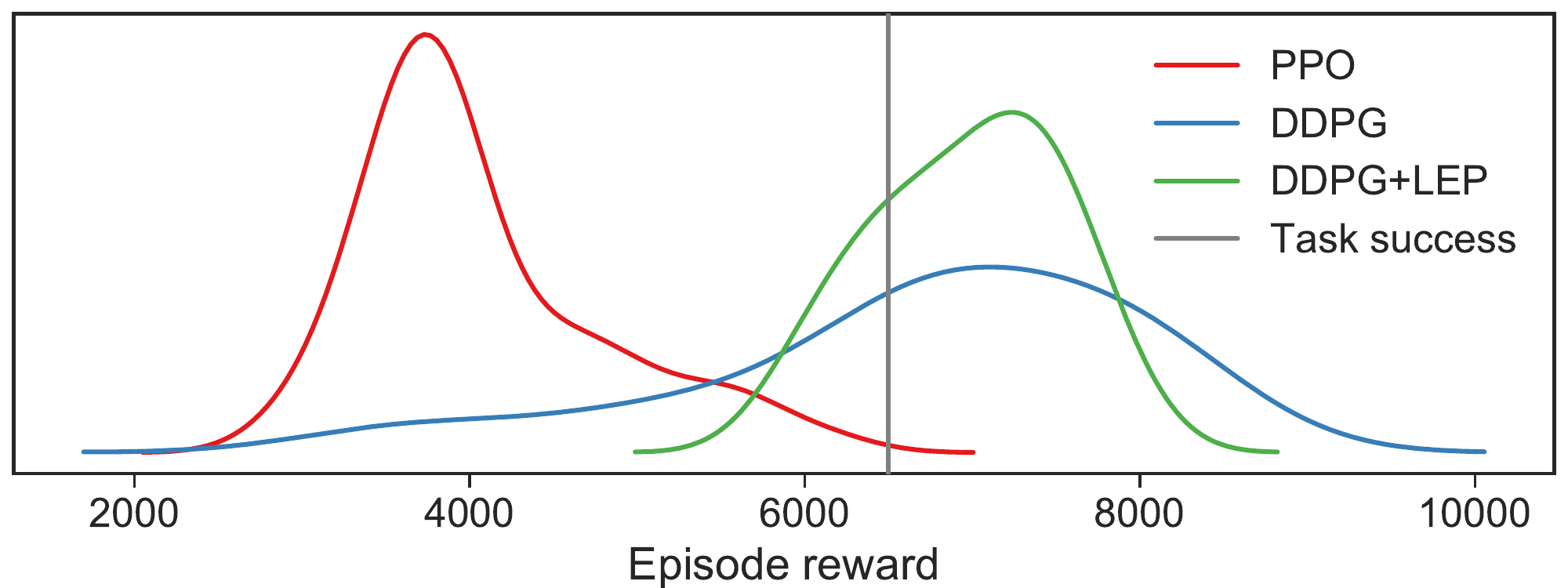}
        \caption{ Learning  results  of  the  best  performing  policies  for
Task 4. (Top) average cumulative reward (bold lines) and variance
(shaded area)  across all  task instances. (Bottom) Final reward distribution.}
        \label{fig:exp3_best}
\end{figure}
\begin{figure}
        \includegraphics[width=\linewidth]{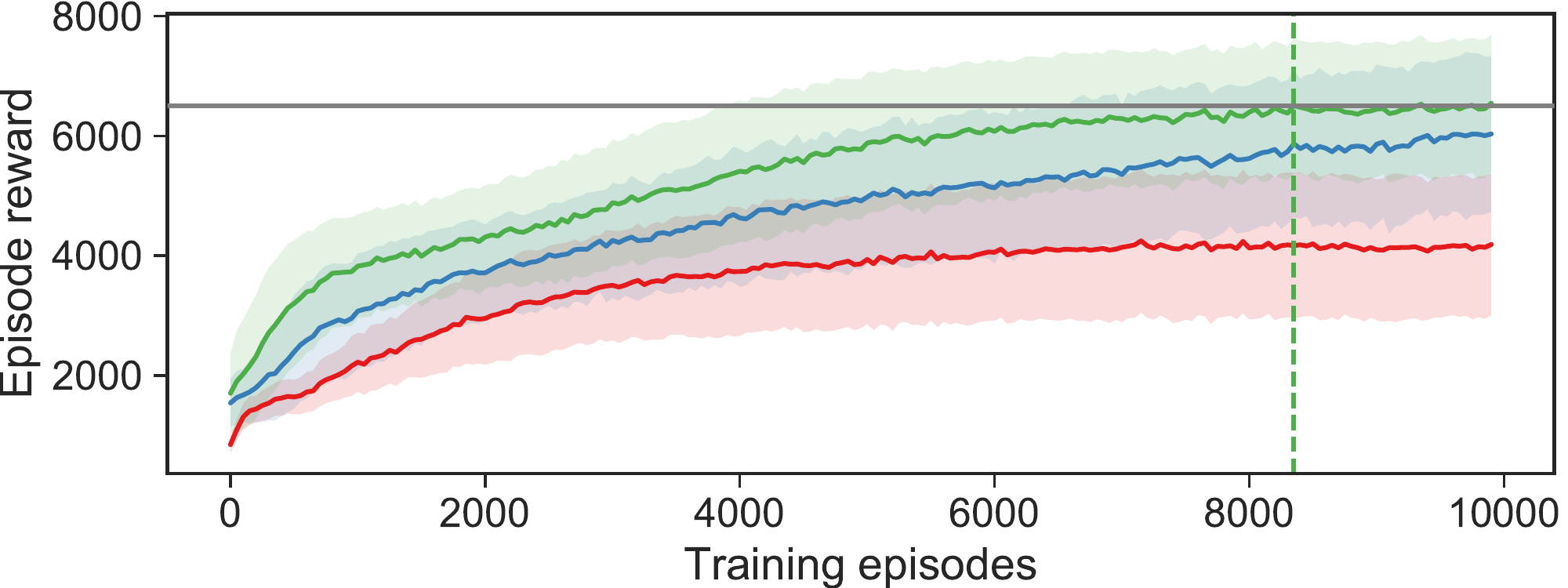}
        \includegraphics[width=\linewidth]{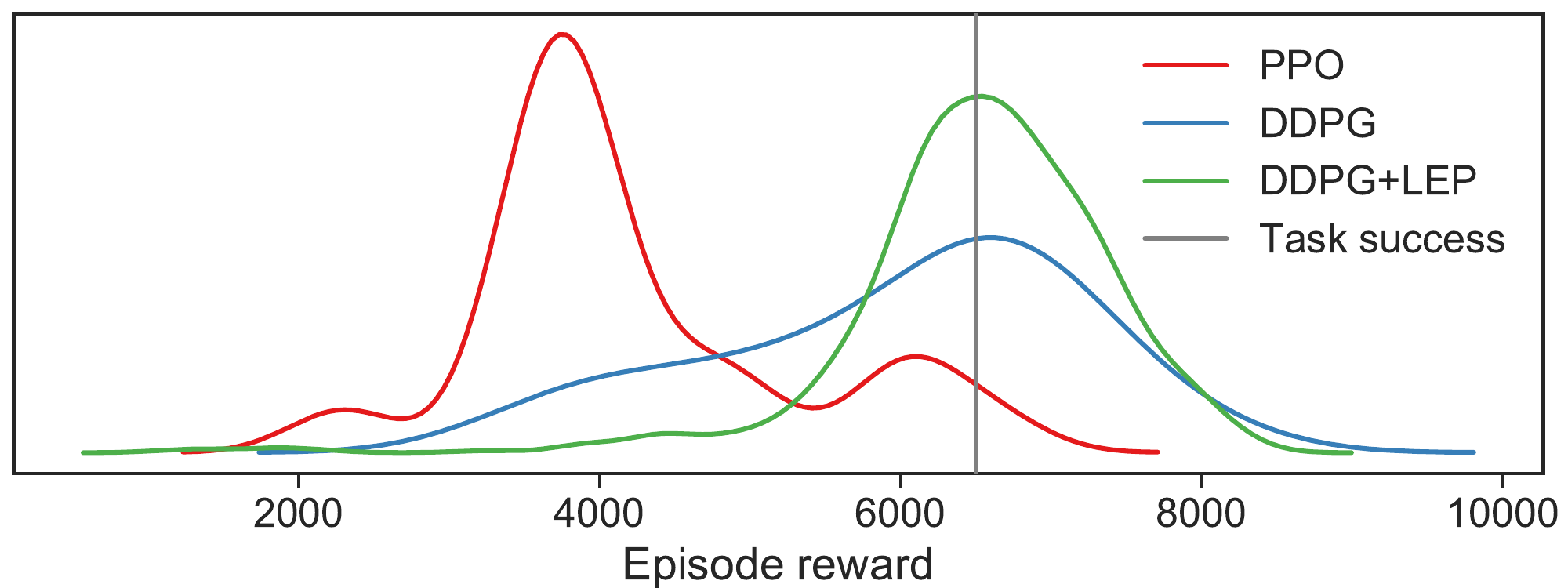}
\caption{Robustness to parametrization results for
Task 4. (Top) average cumulative reward (bold lines) and variance
(shaded area)  across all task instances and all parametrizations. (Bottom) Final reward distribution.}
\label{fig:exp3_dist}
\end{figure}
First, we compare our approach with other algorithms in the same way as before. We still use the best performing  parametrizations of each algorithm to compare the results and in this case use 50 different task instances for the comparison (Figure \ref{fig:exp3_best}).
With the table interaction aspect again present, PPO fails to find satisfactory solutions.
Both standard DDPG and the one using our exploration process do converge, but with our method doing so almost twice as fast.
In addition to the learning curves, Figure \ref{fig:exp3_best} 
also shows the full distributions of the cumulative rewards for the policies for all three algorithms at the end of learning.
This figure clearly suggests that our approach both has a higher percentage of satisfactory solutions as well as practically no policies with really bad scores. This is in contrast with PPO which finds mostly poorly performing policies and DDPG that has a more elongated distribution of results: for certain task instances
it finds solutions but fails to do so for others. These results
additionally support that our approach can speed-up learning 
but more importantly, it suggests that learning performance becomes more consistent and repeatable across task instances.

\subsection{Robustness}
Finally, we would like to demonstrate robustness and the lack of need for tuning parameters in our approach. This is especially important because
oftentimes reinforcement learning algorithms are very sensitive to
parameter tuning and the same experiments with random initial conditions for the system can lead to very different results \cite{colas2018many}.

In all previous experiments, we presented results using the best parametrization for each algorithm (between different noise parameters for standard DDPG and different subsequence lengths used for training the LEP). Now we present results for learning Task 4 for all the parametrization we tested, without making any such choices. The results are presented in Figure \ref{fig:exp3_dist}.
The difference between each algorithm is very clear.
The results for our method are only slightly worse than those we presented for the best parameter configuration, demonstrating its 
robustness to parametrization.
The average policy performance at the end of the training especially is only slightly affected.
Without any tuning our method still produces a majority of satisfactory policies for this complex task.
That is not the case with either standard DDPG or PPO. DDPG becomes significantly worse in this evaluation, with the average not reaching satisfactory value at the completion of the training. PPO was already not performing the tasks with the best parametrization.
These results support the idea that our exploration strategy can not only speed-up learning, but also improve the robustness of the algorithms to parameter tuning, which can be a significant gain
when deploying such algorithms on novel tasks and robots.

\section{DISCUSSION}

The goal of learning the exploration process is to make the behaviors needed to solve a new task more likely to occur during exploration. 
This is done in an effort to speed up learning, as well as to achieve complex behaviors that might otherwise be missed.
In the absence of exploration capable of doing so, we are usually forced to add guiding terms to reward functions to lead the policies in desired direction during training.
Such terms not only require additional tuning, but are also not representative of actual aspects we would like to achieve.
As tasks become more and more complex, with many-part reward functions, the process of adding this guiding information becomes untenable.
This is why we aim to relieve some of this effort by having a good exploration process, freeing up the reward function to just encode the task at hand.

Some of the main questions to be considered when using an approach like the one we present in this work are related to ways in which data from one task can be useful in solving a different one. 
Here, that is reflected in the choices we make in selecting data to train the exploration process, as well as more generally how we structure a curriculum of tasks with a goal of generating more and more complex behaviors on the system.

First, it is worth pointing out that using all the data we have access to, even if it is extensive and varied, might not necessarily be a bad idea.
Barring issues with the model not being able to fit the data correctly, the only downside would be that our model encodes a wider distribution, covering behaviors that might not be directly useful for the current task.
In that case, some of the exploratory behavior might not be relevant, but that should not prevent us from gaining benefits from the rest of it.

In the same way having too much data might not cause issues, lacking data for some part of the behavior needed to solve the task might not be detrimental either.
Using an exploration model trained only on behavior needed for one aspect of the task will not necessarily prevent us from learning how to solve all the other aspects as well.
For example, as can be seen in Figure \ref{fig:exp2}, using an exploration process trained only using free space motions causes no issues in learning on a task where force also needs to be applied.
What is more, learning is significantly faster on the new task than it is with the standard methods.

Taking the two previous points into account we still want to make the best choices we can in building an exploration process with a goal of solving a new task.
For achieving that we should take a look at what kind of motion and interaction behavior is expected in the new task and choose already known tasks exhibiting some parts of that behavior.
The goal being to decompose the task into as many elements that are already contained in some of the known policies.
The same idea applies when building an entire curriculum of tasks to solve on a system, where we should start from the ones easiest to solve and slowly add new aspects as we progress in generating more and more complex behaviors.

\section{CONCLUSION}

In this paper, we presented a novel approach to learn an exploration process for reinforcement learning using previously learned tasks.
The system is built such that as novel tasks are learned, the exploration model can be improved and facilitate learning more complex tasks. This is particularly useful for robotics problems where such hierarchy of tasks (from simple to complex) naturally arises. In our future work we intend to demonstrate the learned policies on a real robot and extend the approach to more complex manipulation tasks.

\bibliographystyle{bibliography/IEEEtran.bst} 
\bibliography{bibliography/IEEEfull,bibliography/references}

\end{document}